%% file: arxiv.tex
\documentclass[10pt,twocolumn,letterpaper]{article}

\usepackage[pagenumbers]{cvpr} 

\makeatletter
\@namedef{ver@everyshi.sty}{}
\makeatother
\usepackage{graphicx}
\usepackage{algorithm}
\usepackage{algorithmic}
\usepackage{amsmath}
\usepackage{amssymb}
\usepackage{booktabs}
\usepackage{color}
\usepackage{xcolor}
\usepackage{multicol}
\usepackage{multirow}
\usepackage{tikz}
\usepackage{enumitem}
\usepackage{pifont}

\usepackage[pagebackref,breaklinks,colorlinks, bookmarks=false]{hyperref}

\usepackage[capitalize]{cleveref}
\crefname{section}{Sec.}{Secs.}
\Crefname{section}{Section}{Sections}
\Crefname{table}{Table}{Tables}
\crefname{table}{Tab.}{Tabs.}
\newcommand{\cmark}{\ding{51}}%
\newcommand{\xmark}{\ding{55}}%
\definecolor{Gray}{gray}{0.5}

\def\name{PF-Track}
\newcommand\mypar[1]{\par\vspace{2.5mm}\noindent\textbf{#1}\;\;}
\input{def}

\begin{document}

\title{Standing Between Past and Future: Spatio-Temporal Modeling for \\ Multi-Camera 3D Multi-Object Tracking}

\author{Ziqi Pang$^{1}$\thanks{Work done while interning at Toyota Research Institute.}, Jie Li$^2$, Pavel Tokmakov$^2$, Dian Chen$^2$, Sergey Zagoruyko$^3$, Yu-Xiong Wang$^1$\thanks{{Corresponding to Ziqi Pang at \texttt{ziqip2@illinois.edu} and Yu-Xiong Wang at \texttt{yxw@illinois.edu}.}} \\
University of Illinois Urbana-Champaign$^1$, Toyota Research Institute$^2$, Woven Planet Level-5$^3$}
\maketitle

\begin{abstract}

This work proposes an end-to-end multi-camera 3D multi-object tracking (MOT) framework. It emphasizes spatio-temporal continuity and integrates both past and future reasoning for tracked objects. Thus, we name it ``Past-and-Future reasoning for Tracking'' (\name). Specifically, our method adopts the ``tracking by attention'' framework and represents tracked instances coherently over time with object queries. To explicitly use historical cues, our ``Past Reasoning'' module learns to refine the tracks and enhance the object features by cross-attending to queries from previous frames and other objects. The ``Future Reasoning'' module digests historical information and predicts robust future trajectories. In the case of long-term occlusions, our method maintains the object positions and enables re-association by integrating motion predictions. On the nuScenes dataset, our method improves AMOTA by a large margin and remarkably reduces ID-Switches by 90\% compared to prior approaches, which is an order of magnitude less. The code and models are made available at \href{https://github.com/TRI-ML/PF-Track}{https://github.com/TRI-ML/PF-Track}.
   
\end{abstract}


\input{sections/1-intro.tex}
\input{sections/2-related.tex}
\input{sections/3-method.tex}
\input{sections/4-experiments}
\input{sections/5-conclusions}

\section*{Appendix}

Our appendix describes the additional experimental analysis and implementation details. The catalog is as below:
\begin{enumerate}[leftmargin=*, noitemsep, nolistsep, label=(\Alph*)]
    \item \textbf{Video demo.} We provide a demo for multi-camera 3D multi-object tracking (MOT) as explained in Sec.~\ref{sec:supp_video}.
    \item \textbf{Implementation details.} We explain the detailed model architecture, procedures for training and inference, and settings for ablation studies in Sec.~\ref{sec:supp_implementation_details}.
    \item \textbf{Additional ablations.} We provide more analysis and experimental results in Sec.~\ref{sec:supp_ablation}.
    \item \textbf{Performance Verification.} For checking the results, we provide the screenshot of the test split results for verification in Sec.~\ref{sec:supp_test_screen_shot}.
\end{enumerate}

\renewcommand\thesection{\Alph{section}}
\renewcommand\thetable{\Alph{table}}
\renewcommand\thefigure{\Alph{figure}}
\renewcommand\thealgorithm{\Alph{algorithm}}
\setcounter{section}{0}
\setcounter{table}{0}
\setcounter{figure}{0}

\section{Multi-camera Tracking Video Demo}
\label{sec:supp_video}

Our demo video is at \url{https://youtu.be/eJghONb2AGg}. It contains:
\begin{itemize}[leftmargin=*, noitemsep, nolistsep]
\item Visualization of 3D MOT results on both surrounding images and Bird's-eye-view.
\item Illustration for addressing occlusions.
\item Qualitative results for predicted trajectories.
\end{itemize}
 
\input{supp_sections/implementation.tex}
\input{supp_sections/more_ablation.tex}
\input{supp_sections/detr3d.tex}
\input{supp_sections/miscellaneous.tex}

{\small
\bibliographystyle{ieee_fullname}
\bibliography{egbib}
}

\end{document}

%% file: def.tex
\usepackage{amsthm, amsmath, amssymb, amsthm} 

\theoremstyle{plain}

\theoremstyle{definition}

\newcommand{\Lcal}{\mathcal{L}}

%% file: sections/1-intro.tex
\section{Introduction}
Reasoning about object trajectories in 3D is the cornerstone of autonomous navigation. While many LiDAR-based approaches exist~\cite{simpletrack,zaech2022learnable,weng2020gnn3dmot}, their applicability is limited by the cost and reliability of the sensor. Detecting, tracking, and forecasting object trajectories only with cameras is hence a critical problem. Significant progress has been achieved on these tasks separately, but they have been historically primarily studied in isolation and combined into a full-stack pipeline in an ad-hoc fashion.

In particular, 3D detection has attracted a lot of attention~\cite{bevformer, detr3d, petr, huang2021bevdet, li2022bevdepth}, but associating these detections over time has been mostly done independently from localization~\cite{scheidegger2018mono,luiten2020track,hu2022monocular}. Recently, a few approaches to end-to-end detection and tracking have been proposed, but they operate on neighboring frames and fail to integrate longer-term spatio-temporal cues~\cite{mutr3d, cc3dt, marinello2022triplettrack, chaabane2021deft}. In the prediction literature, on the other hand, it is common to assume the availability of ground truth object trajectories and HD-Maps~\cite{chang2019argoverse, nuscenes, wilson2021argoverse, ettinger2021large}. A few attempts for a more realistic evaluation have been made~\cite{ivanovic2021heterogeneous,gu2022vip3d}, focusing only on the prediction performance.
\begin{figure}
    \centering
    \includegraphics[width=0.95\linewidth]{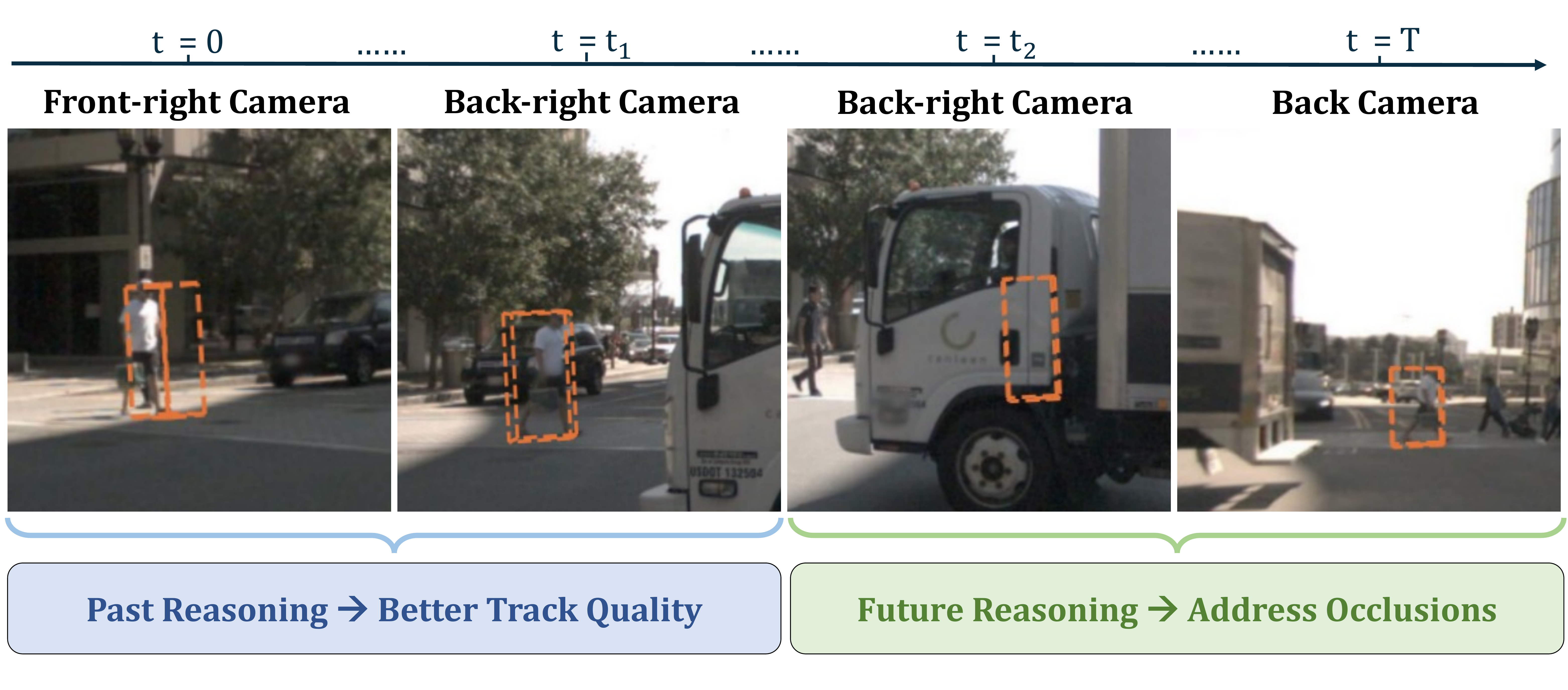}
    \vspace{-2mm}
    \caption{ We visualize the output of our model by projecting predicted 3D bounding boxes onto images. In the beginning, image-based detection can be inaccurate ($t=0$) due to depth ambiguity. With ``Past Reasoning,'' the bounding box quality ($t=t_1$) gradually improves by leveraging historical information. With ``Future Reasoning,'' our \name\ predicts the long-term motions of objects and maintains their states even under occlusions ($t=t_2$) and camera switches. This enables re-association without explicit re-identification ($t=T$), as the object ID does not switch. Our \name\ further combines past and future reasoning in a joint framework to improve spatio-temporal coherence. 
    }
    \vspace{-4mm}
    \label{fig:intro}
\end{figure}

In this paper, we argue that multi-object tracking can be dramatically improved by jointly optimizing the detection-tracking-prediction pipeline, especially in a camera-based system. We provide an intuitive example from our real-world experiment in Fig.~\ref{fig:intro}. At first, the pedestrian is fully visible, but a model with only single-frame information makes a prediction with large deviation (frame $t=0$ in Fig.~\ref{fig:intro}). After this, integrating the temporal information from the past gradually corrects the error over time (frame $t=t_1$ in Fig.~\ref{fig:intro}), by capitalizing on the notion of spatio-temporal continuity. Moreover, as the pedestrian becomes fully occluded (frame $t=t_2$ in Fig.~\ref{fig:intro}), we can still predict their location by using the aggregated past information to estimate a future trajectory. Finally, we can successfully track the pedestrian on re-appearance even on a different camera via long-term prediction, resulting in correct re-association (frame $t=T$ in Fig.~\ref{fig:intro}). The above robust spatio-temporal reasoning is enabled by seamless, bi-directional integration of past and future information, which starkly contrasts with the mainstream pipelines for vision-based, multi-camera, 3D multi-object tracking (3D MOT).

To this end, we propose an end-to-end framework for joint 3D object detection, tracking, and trajectory prediction for the task of 3D MOT, as shown in Fig.~\ref{fig:framework}, adopting the ``tracking by attention''~\cite{trackformer, motr, mutr3d} paradigm. Compared to our closest baseline under the same paradigm~\cite{mutr3d}, we are different in explicit past and future reasoning: a 3D object query consistently represents the object over time, propagates the spatio-temporal information of the object across frames, and generates the corresponding bounding boxes and future trajectories. To exploit spatio-temporal cues, our algorithm leverages simple attention operations to capture object dynamics and interactions, which are then used for track refinement and robust, long-term trajectory prediction. Finally, we close the loop by integrating predicted trajectories back into the tracking module to replace missing detections (\textit{e.g.}, due to an occlusion). To highlight the capability of joint past and future reasoning, our method is named ``Past-and-Future reasoning for Tracking'' (\name).

We provide a comprehensive evaluation of \name\ on nuScenes~\cite{nuscenes} and demonstrate that joint modeling of past and future information provides clear benefits for object tracking. In particular, \name\ decreases ID-Switches by over \textbf{90}\% compared to previous multi-camera 3D MOT methods.

To summarize, our contributions are as follows. 

\begin{enumerate}[leftmargin=*, noitemsep, nolistsep]
    \item We propose an end-to-end vision-only 3D MOT framework that utilizes object-level spatio-temporal reasoning for both past and future information.
    \item Our framework improves the quality of tracks by cross-attending to features from the ``\textbf{past}.''
    \item We propose a joint tracking and prediction pipeline, whose constituent part is ``Future Reasoning'', and demonstrate that  tracking can explicitly benefit from long-term prediction into the ``\textbf{future}.''
    \item Our method establishes new state-of-the-art on large-scale nuScenes dataset~\cite{nuscenes} with significant improvement for both AMOTA and ID-Switch. 
\end{enumerate}

%% file: sections/2-related.tex
\begin{figure*}
    \centering
    \includegraphics[width=0.85\linewidth]{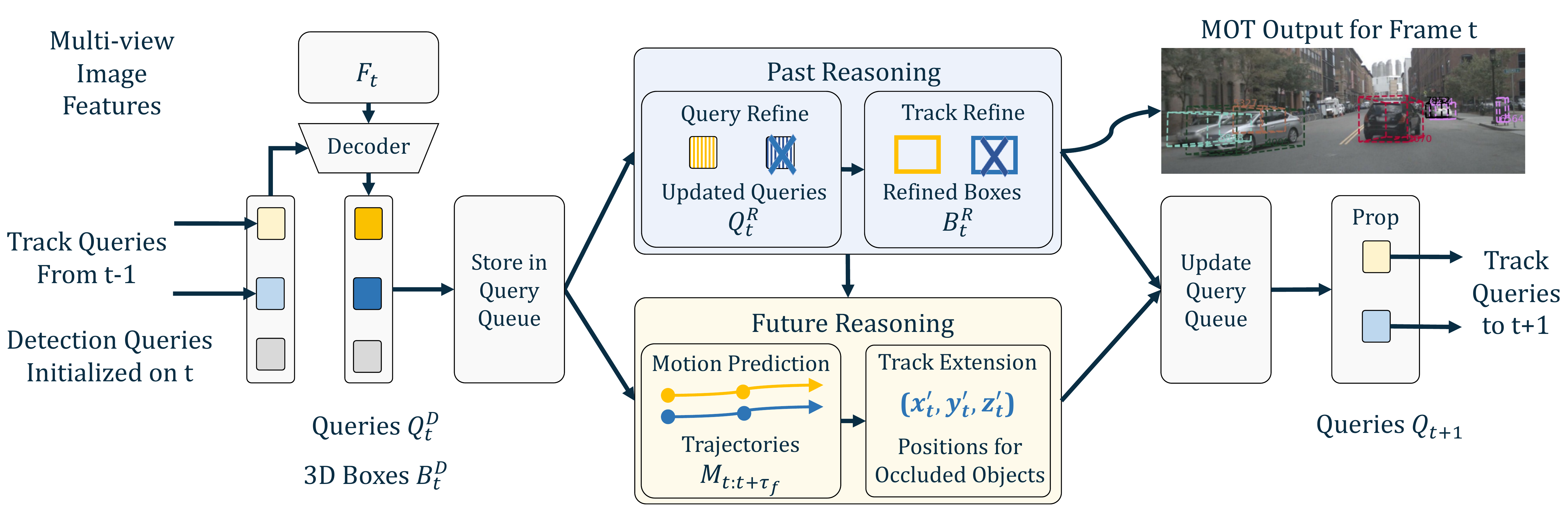}
    \vspace{-3mm}
    \caption{\textbf{\name\ Framework.} \name\ represents objects as queries, decodes image features, and predicts bounding boxes. To improve spatio-temporal coherence, we incorporate novel ``Past Reasoning'' and ``Future Reasoning'' modules. (1) ``Past Reasoning'' refines the features of queries and bounding boxes of tracks by exploiting the historical information in the query queue. (2) ``Future Reasoning'' improves the propagation of queries across frames by estimating long-term future trajectories. Furthermore, if an object is lost due to low confidence or occlusion (blue squares with $\times$), the ``track extension'' module can use a long-term trajectory to maintain its location. Finally, \name\ incorporates past and future reasoning jointly for 3D MOT.  (Best viewed in color, details in Sec.~\ref{sec:pipeline}.)}
    \vspace{-4mm}
    \label{fig:framework}
\end{figure*}

\section{Related Work}

\mypar{LiDAR-based 3D MOT.}
The majority of prior works in 3D MOT leverage the LiDAR modality.
Due to the recent advances in LiDAR-based 3D detection~\cite{yin2021center, pointpillars}, especially the reliable range information, most state-of-the-art 3D MOT algorithms adopt a ``tracking-by-detection'' paradigm~\cite{ab3dmot}. Given single frame detection outputs, different approaches have been proposed to improve data association~\cite{simpletrack,zaech2022learnable,weng2020gnn3dmot}, motion propagation~\cite{chiu2021probabilistic,centertrack}, and life cycling~\cite{simpletrack,wang2021immortal}.
However, most of these works assume the localization accuracy of detection output. Therefore, data association is usually conducted based on location, optionally combined with abstracted object attributes (\textit{e.g.}, 3D intersection over untion (3D IoU)~\cite{ab3dmot}, 3D generalized intersection over union (GIoU)~\cite{simpletrack}, and L2 distance~\cite{yin2021center}). This bias causes the proposed systems to be fragile when migrated into the camera modality, where 3D detection suffers from higher localization uncertainty. Although the latest methods incorporate learning-based algorithms to improve association with high-fidelity features such as low-level features from point clouds~\cite{stearns2022spot} or intermediate features from cameras~\cite{chiu2021probabilistic}, these approaches are built on top of the LiDAR-based frameworks and share their dependence on localization quality. 

\mypar{Camera-based 2D MOT.}
Camera-based multi-object tracking in 2D is a classic task in computer vision. Dominated by ``tracking by detection'' paradigm~\cite{bewley2016simple}, 2D MOT has seen more success in leveraging high-fidelity features~\cite{wojke2017simple,tang2017multiple,zhang2021fairmot,peng2021transmot}. 
Earlier works like DeepSORT~\cite{wojke2017simple} leveraged intermediate features from a deep net to measure appearance similarity.
FairMOT~\cite{zhang2021fairmot} employed an additional Re-ID branch to learn discriminative features in a detection network. 
TransMOT~\cite{peng2021transmot} proposed to incorporate spatio-temporal features using a graph network. 

\mypar{Camera-based 3D MOT.}
Camera-based 3D MOT has recently drawn more attention in autonomous driving applications thanks to advances in monocular depth estimation~\cite{dorncvpr,packnet,godard2019digging} and image-based 3D object detection~\cite{CaDDN,park2021dd3d, bevformer, petrv2, li2022bevdepth, huang2021bevdet, petr, detr3d, wang2021fcos3d}. Early methods adapt the 2D MOT algorithms and lift the 2D tracking result using monocular depth~\cite{centertrack, tokmakov2021learning}.
More recent approaches employ additional 3D information in data association~\cite{scheidegger2018mono,luiten2020track,hu2022monocular}. \cite{luiten2020track} proposes to leverage 3D reconstruction, and \cite{hu2022monocular} augments the 2D Re-ID features with 3D attributes (\textit{e.g.} depth and orientation).
CC-3DT~\cite{cc3dt} merges the multi-view camera features for identical objects to improve the cross-time cross-view association.
However, considering or correcting the high uncertainty and bias in camera-based 3D detection has been less explored. In this work, we leverage long-term object reasoning, especially past reasoning, to improve the quality of 3D bounding boxes.

\mypar{Tracking by Attention.} A rising trend in MOT is the ``tracking by attention'' paradigm~\cite{motr,sun2020transtrack,trackformer,mutr3d}, inspired by the novel transformer-based detection architecture DETR~\cite{carion2020end}. MOTR~\cite{motr} and Trackformer~\cite{trackformer} extended the query-based detection framework in DETR~\cite{carion2020end} by propagating queries across different frames. In this paradigm, the data association is replaced by ``detection'' in the current frame with a set of track queries. MUTR3D~\cite{mutr3d} proposes the first framework applying this paradigm to the 3D MOT domain. It uses a 3D track query to jointly model object features across timestamps and multi-views. Despite its improvement at the time, MUTR3D mostly follows the designs of 2D MOT methods and does not include special treatment to improve the localization quality of tracks and better propagate the queries to future frames. Our proposed algorithm also operates in the  ``tracking by attention'' paradigm but extends the temporal horizon of existing methods. In particular, we demonstrate that joint past and future reasoning can improve the tracking framework by providing a strong spatio-temporal object representation.

\mypar{Motion Prediction.} Predicting agent trajectories is critical for self-driving~\cite{ngiam2021scene, gao2020vectornet, yuan2021agentformer, ivanovic2019trajectron, salzmann2020trajectron++, liu2021multimodal, shi2022motion}. The most common setting is to predict from clean tracks annotated by humans or auto-labeling~\cite{sun2020scalability, ettinger2021large, chang2019argoverse, wilson2021argoverse}. Numerous studies focus on end-to-end prediction from perception~\cite{weng2022mtp, weng2022whose, weng2021inverting, fiery, beverse, casas2021mp3, luo2018fast, phillips2021deep, shah2020liranet, liang2020pnpnet, peri2022forecasting, akan2022stretchbev}, especially how to improve motion prediction directly from perception. However, our objective is different: Could a motion prediction model benefit 3D MOT? In the 2D setting, this problem has received only limited attention recently~\cite{dendorfer2022quo}. Our algorithm advances this research into a more challenging multi-camera, 3D scenario and does not require explicit re-identification.

%% file: sections/3-method.tex
\section{Method: \name}
This section introduces our novel 3D multi-object tracking framework, shown in Fig.~\ref{fig:framework}. It is centered around explicit past and future modeling of object trajectories in an end-to-end framework. We first provide an overview of the pipeline in Sec.~\ref{sec:pipeline}, and then explain how to efficiently leverage ``Past'' (Sec.~\ref{sec:past}) and ``Future'' (Sec.~\ref{sec:future}) information. Finally, we summarize the losses used in our framework in Sec.~\ref{sec:loss}.

\subsection{\name\ Pipeline}~\label{sec:pipeline}
Our proposed \name\ iteratively uses a set of object queries~\cite{mutr3d,motr,trackformer} to tackle multi-view, multi-object, 3D tracking. At each timestamp $t$, given $K$ images $\textbf{I}_t^{k}$ from surrounding cameras, the objective of 3D MOT is to generate object detections with consistent IDs across frames, denoted by $\textbf{B}_t=\{\textbf{b}_t^i\}$, where $i$ is an object ID. 

\mypar{3D Object Queries.} The entry point in our framework is to receive the object queries $\textbf{Q}_t=\{\textbf{q}^i_t\}$ propagated from the previous frame $t-1$ (yellow and blue squares in Fig.~\ref{fig:framework}), which represent the tracked objects:
\begin{equation}
    \textbf{Q}_t\leftarrow \mathbf{Prop}(\textbf{Q}_{t-1}).
    \label{eq:prop}
\end{equation}
Such a query-based design naturally addresses the task of tracking as the queries carry the identity of objects over time. Apart from queries from the previous frame that represent tracked instances, we also add a fixed number of detection queries (gray squares in Fig.~\ref{fig:framework}) to discover new objects. In practice, we use 500 detection queries initialized as learnable embeddings. 

Each query $\textbf{q}_t^i\in \textbf{Q}_t$ represents a unique 3D object with a feature vector $\textbf{f}_t^i$ and a 3D location $\textbf{c}_t^i$: $\textbf{q}^i_t = \{\textbf{f}_t^i, \textbf{c}_t^i\}$. Here we highlight that the query position is an active participant in decoding the bounding boxes of objects below.

\mypar{Decoder.} To predict 3D bounding boxes and update queries with the latest image inputs, \name\ adopts an attention-based detection architecture~\cite{carion2020end,zhu2020deformable} to decode image features $\textbf{F}_t$ with object queries:
\begin{equation}
   \textbf{B}^D_t,\textbf{Q}^D_t\leftarrow \mathbf{Decoder}(\textbf{F}_t,\textbf{Q}_t),
   \label{eq:det}
\end{equation}
where $\textbf{B}^D_t$ and $\textbf{Q}^D_t$ are the detected 3D bounding boxes and updated query features, respectively. In the decoding process, the decoder lifts the 3D positions $\textbf{c}_t^i$ of queries into positional embeddings to concentrate on the image regions relevant to the spatial locations of the objects. While the design of \name\ is agnostic to query-based detection algorithms, we mainly adopt a current state-of-the-art 3D detector, PETR~\cite{petr}, for experiments.

\mypar{Past and Future Reasoning for Refinement and Propagation.} 
After decoding the queries and boxes from single-frame image features, \name\ conducts past and future reasoning sequentially to (1) refine the current detections $\textbf{B}_t^D$ into ${\textbf{B}}^R_t$ and queries $\textbf{Q}_t^D$ into $\textbf{Q}^R_t$. ($R$ is short for ``refinement.''); (2) propagate the queries to the next timestamp with the predicted motions.

``Past Reasoning'' $\textbf{PR}(\cdot)$ is the component that aggregates the information from previous frames to generate refined queries $\textbf{Q}^R_t$ and refined bounding boxes ${\textbf{B}}^R_t$:
\begin{equation}
    \label{eq:past_and_future}
    \textbf{Q}^R_t, {\textbf{B}}^R_t \xleftarrow[]{} \textbf{PR}(\textbf{Q}^D_{t}, \textbf{B}^D_t,{\textbf{Q}}_{t-\tau_h:t-1}, ).
\end{equation}
In practice, the historical queries ${\textbf{Q}}_{t-\tau_h:t-1}$ come from a \textbf{query queue} that maintains the queries from past $\tau_h$ frames ($h$ for ``history'').

After past reasoning, the ``Future Reasoning'' module $\textbf{FR}(\cdot)$ improves the coherence of object positions from the aspect of query propagation. It achieves this by forecasting the motions up to $\tau_f$ frames ($f$ for ``future'') and transforms the positions of queries accordingly:
\begin{equation}
   \textbf{Q}_{t+1}, \textbf{M}_{t:t+\tau_f} \xleftarrow[]{} \textbf{FR}(\textbf{Q}^R_t, {\textbf{Q}}_{t-\tau_h:t-1}).
\end{equation}
Specifically, future reasoning extracts the object dynamics from historical query features to predict the trajectories $\textbf{M}_{t:t+\tau_f}$. The single-step movement $\textbf{M}_{t:t+1}$ is leveraged to propagate the current queries $\textbf{Q}^R_t$ to the next timestamp, and long-term trajectories $\textbf{M}_{t+1:t+\tau_f}$ are used for addressing occlusions. The ``Track Extension'' in Fig.~\ref{fig:framework} refers to occlusion reasoning through the predicted trajectories. 

\name\ iteratively executes the above procedures. The refined 3D bounding boxes ${\textbf{B}}^R_t$ are the output for 3D MOT.

\subsection{Past Reasoning}
\label{sec:past}

To address the uncertainty of detection in vision-only 3D localization, past reasoning focuses on two aspects: (1) enhancing the query features by attending to historical embeddings; (2) refining the tracks by adjusting the bounding boxes using the improved query features.

\begin{figure}[tb]
    \centering
    \includegraphics[width=0.99\linewidth]{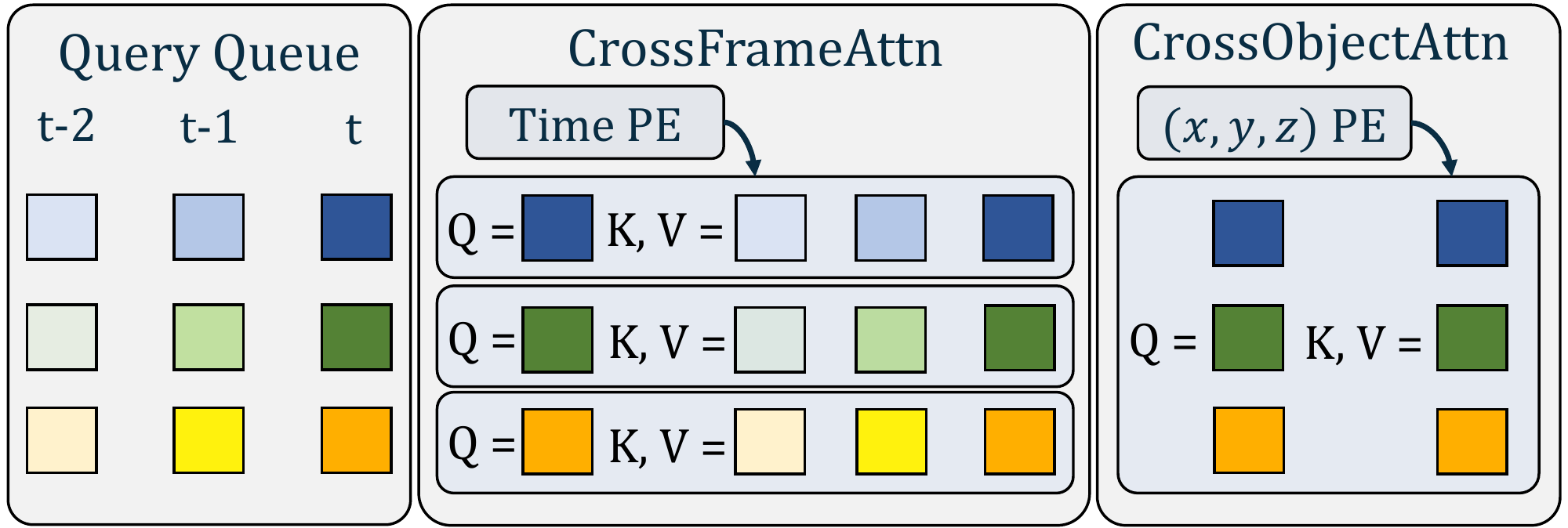}
    \caption{\textbf{Query Refinement.} ``\textit{Cross-frame}'' and ``\textit{Cross-object}'' attention modules process the query queue to capture the temporal and inter-object relationship, respectively. They apply the positional encoding for time $t$ and spatial locations $(x, y, z)$, respectively. (Best viewed in color.)}
    \vspace{-4mm}
    \label{fig:cross_frame_obj}
\end{figure}

\mypar{Query Refinement: from $\textbf{Q}^D_t$ to $\textbf{Q}^R_t$.} 
We first apply attention across the time and instance axes to explicitly update the query features with historical information, as illustrated in Fig.~\ref{fig:cross_frame_obj}.
{``Cross-frame'' attention} encourages the interplay of features within a history window of $\tau_h$ frames per object:
\begin{equation}
\label{eq:cross_frame_attn}
\begin{split}
     {\textbf{f}}^i_{t} \xleftarrow[]{} \mathbf{CrossFrameAttn}(& \text{Q}=\textbf{f}^i_{t}, \\ &\text{K}=\textbf{f}^{i}_{t-\tau_h:t}, \text{V}=\textbf{f}^{i}_{t-\tau_h:t}, \\ & \text{PE}=\mathbf{Pos}(t-\tau_h:t)),
\end{split}
\end{equation}
where, $\mathbf{Pos}(t-\tau_h:t)$ converts the timestamps into positional embedding, and the history frames with empty features are ignored for attention computation.
 
Then past reasoning applies ``cross-object'' attention to incorporate the context information and encourage more discriminative feature representation for each object. 
In particular, cross-object attention (Fig.~\ref{fig:cross_frame_obj}, right) further updates the query features via 
\begin{equation}
\label{eq:cross_object_attn}
\begin{split}
     \textbf{f}^{1:N_t}_{t} \xleftarrow[]{}
     \mathbf{CrossObjectAttn}(& \text{Q,K,V}=\textbf{f}^{1:N_t}_{t}, \\ & \text{PE}=\mathbf{Pos}(\textbf{c}^{1:N_t})),
\end{split}
\end{equation}
where cross-object attention exchanges the features of $N_t$ objects guided by their 3D positional embedding $\mathbf{Pos}(\textbf{c}^{1:N_t})$. The final output $\textbf{f}^{1:N_t}_{t}$ becomes the refined feature vectors in queries $\mathbf{Q}^R_t$.

As a brief remark, decoupling cross-frame and cross-object attention exhibits two advantages. Firstly, separating attention across frames (cross-frame) and objects (cross-object) enables us to design specialized positional encoding of time and locations for each of them. Secondly, it decreases the computational complexity from $\mathcal{O}(N_t^2\tau_h^2)$ for the global cross-attention to $\mathcal{O}(N_t^2+N_t\tau_b^2)$, which is significantly less. Our design is also closely related to how motion prediction methods~\cite{ngiam2021scene, gao2020vectornet} model spatial-temporal relationships. 

\mypar{Track Refinement: from $\mathbf{B}^D_t$ to $\mathbf{B}^R_t$.} With the queries refined by historical information, past reasoning further uses track refinement to improve the 3D bounding box quality. As specified in Eqn.~\ref{eq:xyz_refinement}, we apply a multi-layer perceptron (MLP) to predict the updated properties of objects, including center residuals ($\Delta {x}, \Delta {y}, \Delta {z}$), size ($l,w,h$), orientations ($\theta$), velocities ($\textbf{v}$), and scores ($s$):
\begin{equation}
\label{eq:xyz_refinement}
     (\Delta {x}, \Delta {y}, \Delta {z}, {l}, {w}, {h}, {\theta}, {\mathbf{v}}, s)^i = \mathbf{MLP}(\textbf{f}_t^i).
\end{equation}
These are then used to adjust the original boxes as follows:
\begin{align}
 \textbf{b}_t^i = (\Delta {x}+x_t^i, \Delta {y}+ y_t^i, \Delta {z}+z_t^i, {l}, {w}, {h}, {\theta}, {\mathbf{v}}, s),
\end{align}
resulting in $\textbf{B}^R_t$, which is the final model output at frame $t$.

\subsection{Future Reasoning}
\label{sec:future}
``Future Reasoning'' concentrates on improving the propagation of queries across frames to benefit spatio-temporal coherence. It first learns a trajectory prediction, which is used for moving queries across adjacent frames. Then future reasoning exploits the predicted long-term trajectories for maintaining the positions of occluded or noisy tracks.

\mypar{Motion Prediction.} Trajectory prediction supervises the model's ability to capture object movements and is further beneficial for propagating query positions across timestamps. Similar to past reasoning, our future reasoning model adopts a simple attention-based architecture. Firstly, we generate the motion embeddings for $\tau_f$ timestamps $\textbf{mf}^{i}_{t:t+\tau_f}$ with a cross-frame attention:
\begin{equation}
\begin{split}
    \textbf{mf}^i_{t:t+\tau_f}\! \xleftarrow{}\! \mathbf{Cross}&\mathbf{FrameAttn}( \\
    & \text{Q}=\textbf{mf}^i_{t:t+\tau_f}, \\ &\text{K},\text{V}=\textbf{f}^i_{t-\tau_h:t}, \textbf{f}^i_{t-\tau_h:t}, \\ & \text{PE}=\mathbf{Pos}(t\! -\!\tau_h\! :\! t\! +\! \tau_f)),
\end{split}
\end{equation}
where $\textbf{mf}^i_{t:t+\tau_f}$ are initialized as zeros, and historical features $\textbf{f}^i_{t-\tau_h:t}$ serve as the source of information. Then the movement at every timestamp is decoded by an MLP:
\begin{align}
    \textbf{m}^i_{t:t+\tau_f} = \mathbf{MLP}({\mathbf{mf}}^{i}_{t:T+\tau_f}),
\end{align} 
and the object trajectory in the 3D space can be recovered by combining these frame-level outputs.
Our architecture is inspired by SceneTransformer~\cite{ngiam2021scene}, which also employs a fully-attention-based architecture. 

The predicted trajectories $\textbf{m}^i_{t:t+\tau_f}$ have better fidelity compared to the velocities $\textbf{v}$ predicted by the decoder in $\textbf{B}^R_t$ and $\textbf{B}^D_t$. Thus, we can propagate the positions of queries by adding a single step of the trajectory:
\begin{align}
\textbf{c}^i_{t+1}=\textbf{c}^i_{t}+\textbf{m}^i_{t:t+1}.
\end{align}

\mypar{Track Extension.}
To handle occlusions or noisy observations, we propose to extend the tracks using the predicted trajectories. In particular, we replace missing or low-confidence detections with the output of our motion prediction module, which is initialized from confident observations.
Previous 3D MOT approaches either terminate the tracks or prolong them with heuristic motion models (\textit{e.g.} Kalman filters) under such conditions. However, these solutions both could lead to ID-Switches due to ``early termination''~\cite{simpletrack} or false associations. In contrast, our learnable motion prediction module and track extension strategy are more accurate and robust. 

\begin{figure}
    \centering
    \includegraphics[width=0.80\linewidth]{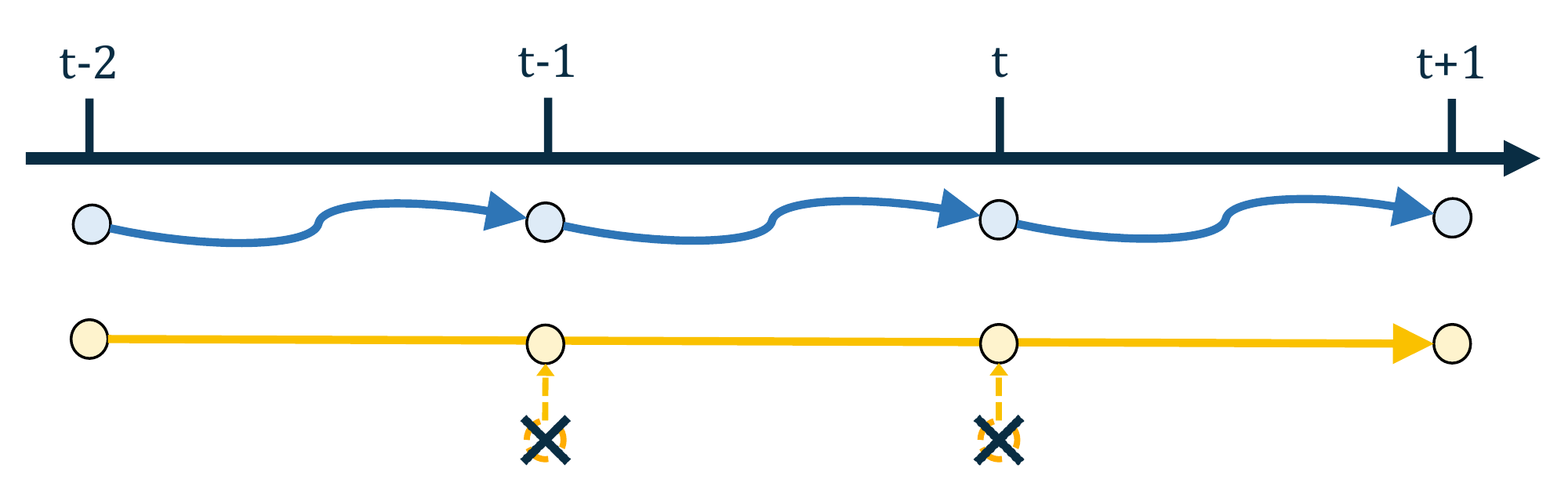}
    \vspace{-2mm}
    \caption{\textbf{Track extension.} \name\ updates object positions and predicts future trajectories at every timestamp (top row). However, if the object cannot be confidently localized (\emph{e.g.} due to occlusion or a noisy observation, bottom row at frames $t-1$ and $t$), our method will rely on the long-term trajectories predicted from confident timestamps (frame $t-2$) to infer the positions of this object and ignore the noisy observations (crossed-out circles).}
    \vspace{-4mm}
    \label{fig:prediction_for_tracking}
\end{figure}

We visualize the high-level intuition in Fig.~\ref{fig:prediction_for_tracking} and provide more details in Sec.~\ref{sup:sec:algo_track_extension} (appendix). In Fig.~\ref{fig:prediction_for_tracking}, the long-term trajectories assist the propagation of the yellow instance (bottom row). When \name\ encounters noisy observation or occlusion cases, it relies on the motion predictions from previous confident frames to simulate the movements of occluded objects. In extreme cases, our model is able to handle occlusion length of $\tau_f-1$ frames. Our ablation study in Sec.~\ref{sec:ablation} demonstrates that track extension can decrease ID-Switch by a large margin. To the best of our knowledge, we are \emph{the first to incorporate long-term prediction into a query-based framework and address occlusion without explicit re-identification}.

\subsection{Loss Functions}
\label{sec:loss}

Our final loss function is defined as follows:
\begin{align}
    \label{eq:losses}
    \begin{split}
\Lcal = & \lambda^D_{\text{cls}}\Lcal^D_{\text{cls}} + \lambda^D_{\text{box}}\Lcal^D_{\text{box}} + \\
    & \lambda^R_{\text{cls}}\Lcal^R_{\text{cls}} + \lambda^{R}_{\text{box}}\Lcal^R_{\text{box}} + \lambda_{f} \Lcal_{f}
    \end{split}
\end{align}
where $\Lcal_{\text{cls}}$ and $\Lcal^R_{\text{cls}}$ are focal loss~\cite{lin2017focal} with the coefficients of $\lambda^D_{\text{cls}}$ and $\lambda^R_{\text{cls}}$. They supervise the classification scores of $\textbf{B}^D_t$ and $\textbf{B}^R_t$, respectively. $\Lcal^D_{\text{box}}$ and $\Lcal^R_{\text{box}}$ are both L1 loss applied to $\textbf{B}^D_t$ and $\textbf{B}^R_t$ for bounding box regression. Their coefficients are $\lambda^D_{\text{box}}$ and $\lambda^R_{\text{box}}$. The motion prediction loss $\Lcal_{f}$ is an L1 loss between the movements of predicted and ground truth trajectories, weighted by $\lambda_f$. The ground truth assignment couples a query with a consistent ground truth instance over time to encourage ID consistency. We discuss more details in Sec.~\ref{sec:supp_implementation_details} (appendix).

%% file: sections/4-experiments.tex
\section{Experiments}

\subsection{Datasets and Metrics}

\mypar{Datasets.} We conduct experiments on the large-scale self-driving dataset nuScenes~\cite{nuscenes}. It contains 1,000 video sequences with multiple modalities, including RGB images from 6 surrounding cameras, and point clouds from LiDAR and Radar. In this paper, we use camera sensors only. Every sequence spans roughly 20 seconds with keyframes annotated at 2Hz. The dataset  provides 1.4M 3D bounding boxes covering 10 types of common objects on the road. For the tracking task, nuScenes selects a subset of 7 mobile categories, such as cars, pedestrians, and motorcycles, and excludes static objects like traffic cones.

\mypar{Metrics.} We strictly follow the official evaluation metrics for multi-object tracking tasks from nuScenes. It modifies CLEAR MOT metrics~\cite{bernardin2008evaluating} by considering multiple recall thresholds. The main metric is ``Average Multi-Object Tracking Accuracy'' (AMOTA)~\cite{ab3dmot}. Meanwhile, we also consider other analytical metrics such as ``Identity Switches'' (IDS) and ``Average Multi-Object Tracking Precision'' (AMOTP).

\input{tables/sota_comparison.tex}

\subsection{Implementation Details} 
\label{sec:implementation_details}

Due to space limits, we clarify two training settings here and describe more implementation detail in Sec.~\ref{sec:supp_implementation_details} (appendix). In our implementation, every training sample contains three adjacent frames from different timestamps. However, it requires extensive computation as every frame contains six high-resolution images. Therefore, we adopt two settings that downsample images to different resolutions, motivated by PETR~\cite{petr}.

\noindent\textbf{Full-resolution.} On every time frame, we crop the raw resolution images, $1600\times900$ to $1600\times 640$, leaving the sky area out. However, training a multi-frame tracker on this resolution would not fit in a single A100 GPU. Thus, we first pretrain the backbone with single-frame detection for 24 epochs, following some previous works~\cite{tokmakov2021learning}. Then we fix the backbone and train the tracker on three-frame samples for another 24 epochs.
We only use this setting for full model results indicated with ``-F'' in Tab.~\ref{tab:sota}.

\noindent\textbf{Small-resolution.} We apply a small-resolution setting for all of our ablation analyses unless specified. In this setting, we downsample the cropped images to a resolution of $800\times 320$. 
We first train a single-frame detection model for 12 epochs and then train the tracker on three-frame samples for another 12 epochs.

\subsection{State-of-the-art Comparison on nuScenes}

In Tab.~\ref{tab:sota}, we compare our model performance with the other published camera-based 3D MOT algorithms on nuScenes. Our approach establishes a new state-of-the-art with significant improvements on every metric. Our AMOTA improves more than $7\%$ on the test set and $12\%$ on the validation set over the previous methods, including a very strong concurrent work~\cite{cc3dt}. 
It is worth noting that with more established tracks (higher recall), \emph{our ID-Switch number is only $10\%$ of previous methods eliminating more than $90\%$ of the ID-switching errors}. This result indicates the strong association ability of our algorithm attributed to leveraging both past and future reasoning. The advantage of our model holds \emph{even when trained in the low-resolution setting}, whereas most of the previous works use full-resolution.  

\subsection{Ablation Studies}
\label{sec:ablation}
\input{tables/full_ablations.tex}
\mypar{Efficacy of Past and Future Reasoning.} In Tab.~\ref{tab:ablation}, we analyze the importance of individual modules for our model's performance using the validation set of nuScenes. In particular, we evaluate the following variants. (1) \textbf{Baseline.} Our baseline is a ``tracking by attention'' model without explicit spatio-temporal reasoning (row 1). It is a strong baseline and outperforms prior work in Tab.~\ref{tab:sota}. (2) \textbf{Past Reasoning.} We first analyze the effect of query refinement (row 2), which explicitly incorporates the historical queries via cross-attention. As illustrated, it improves the overall tracking quality. We then exploit the enhanced feature to refine the 3D bounding boxes of tracks (row 3). It decreases ID-Switch and AMOTP, which indicates that track refinement is useful for 3D MOT. (3) \textbf{Future Reasoning.} Next, we demonstrate that learning to predict object motion and propagate positions (row 4) is beneficial for modeling object dynamics and leads to improved tracking performance. In addition, using the long-term trajectory predictions to replace low-confidence localizations (row 5) results in a 67\% drop in ID-Switches. (4) \textbf{Joint Past and Future Reasoning.} Finally, combining past and future reasoning into an end-to-end framework shown in Fig.~\ref{fig:framework} (row 6) allows our model to achieve top performance. This result confirms that past and future reasoning are mutually beneficial for 3D MOT.

\begin{figure}[t]
    \centering
    \includegraphics[width=1.0\linewidth]{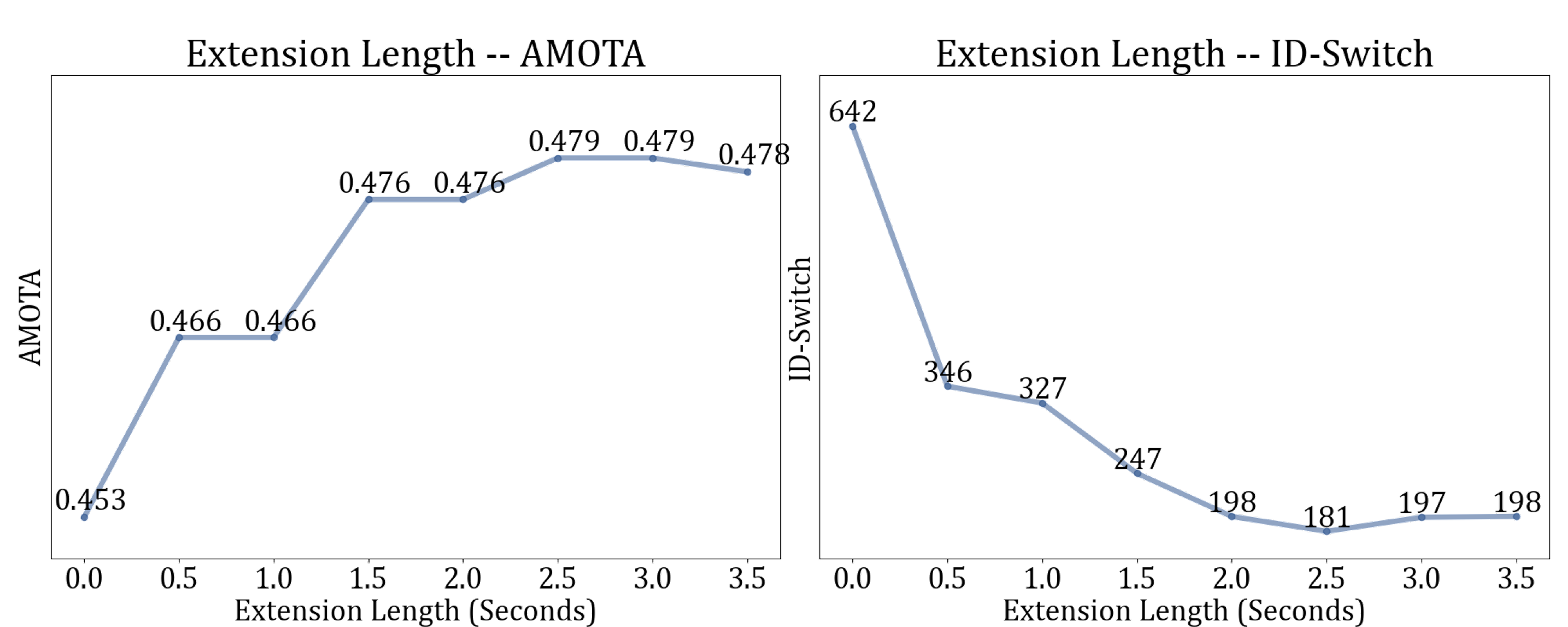}
    \caption{\textbf{Track extension assists MOT.} By using the predicted trajectories to maintain the states for low-confidence tracks, we significantly improve AMOTA and decrease ID-Switch.}
    \vspace{-4mm}
    \label{fig:life_cycle}
\end{figure}

\mypar{Length of Track Extension.} In Fig.~\ref{fig:life_cycle}, we analyze the effect of track extension length on AMOTA and ID-Switches using our best-performing full-resolution model on the validation split of nuScenes. Compared to not using the extension strategy (0.0s), prolonging the tracks strongly improves the performance up to 2 seconds. Then the metrics saturate because only a few objects reappear after such a long period. \emph{Please note that these improvements are achieved without explicit re-identification.}  

\begin{figure*}[tb]
    \centering
    \includegraphics[width=0.80\linewidth]{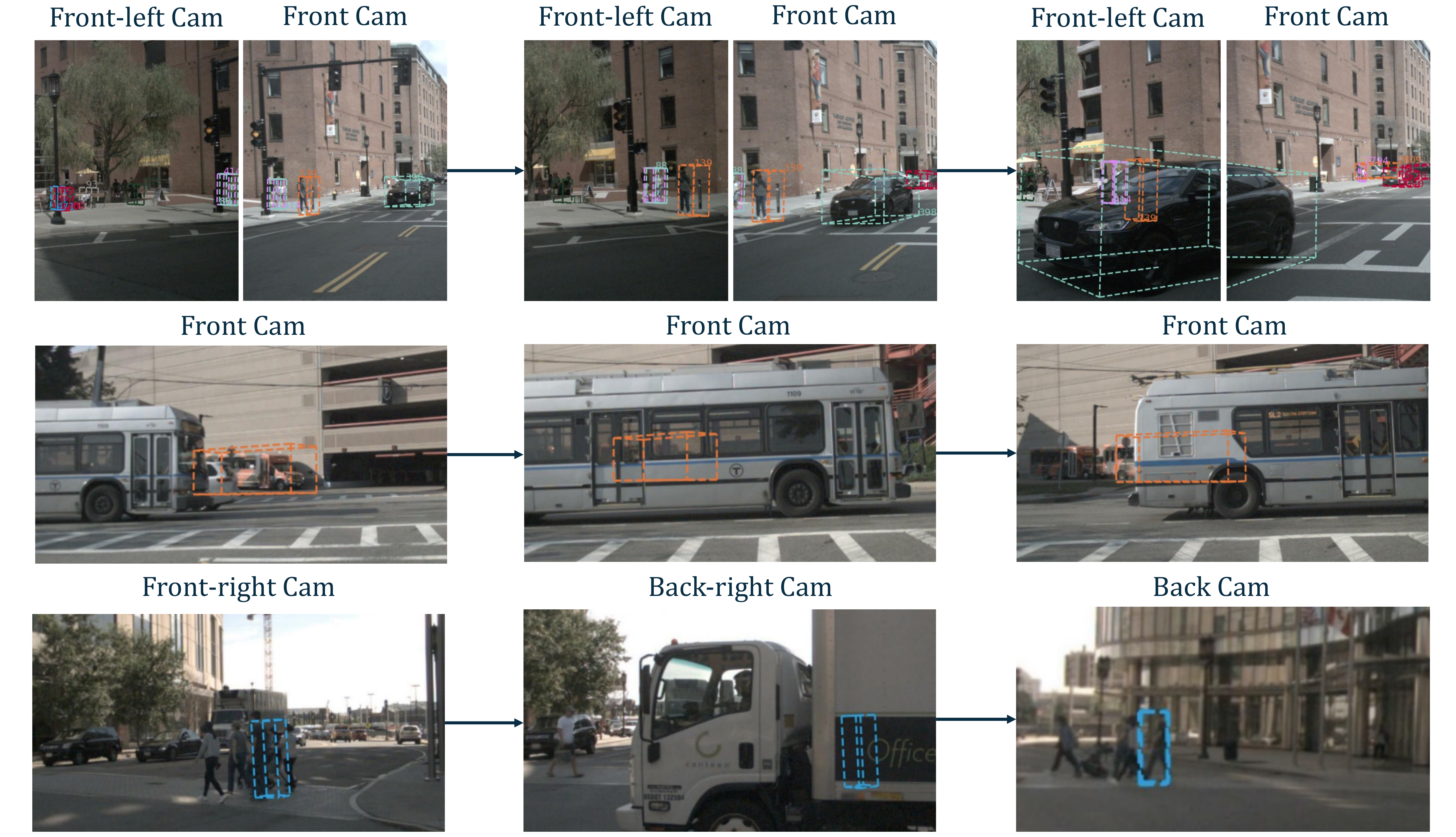}
    \vspace{-2mm}
    \caption{\textbf{Qualitative results for 3D MOT.} (1) In the top row, we provide image-level 3D MOT results. The figures highlight the consistency across images, such as the vehicles crossing the front-left and front cameras. (2) In the middle and bottom rows, we provide two dedicated examples for addressing large and small objects' occlusions. 
    }
    \vspace{-4mm}
    \label{fig:vis_track}
\end{figure*}

\mypar{Length of Prediction in Future Reasoning.} Next, we analyze how the prediction length changes the 3D MOT performance on nuScenes validation split in Tab.~\ref{tab:prediction_length}. Concretely, we train three different full-fledged models with the prediction length of 2.0, 3.0, and 4.0 seconds. AMOTA and ID-Switch indicate that 4.0s (8 frames) has a slight advantage over 2.0s (4 frames) and 3.0s (6 frames). This result indicates that learning trajectory forecasting with longer horizon benefits our 3D MOT framework.

\input{tables/prediction_length.tex}

\mypar{Comparison with ``Tracking by Detection'' Baselines.} In Tab.~\ref{tab:tracking_by_detection}, we compare the performance between our end-to-end framework and previous ``tracking by detection'' algorithms~\cite{ab3dmot, simpletrack, yin2021center}, which are strong baselines for LiDAR-based 3D MOT. For these experiments, we also use the validation split of nuScenes. For a fair comparison, we evaluate these methods with PETR~\cite{petr} detections and tune their hyper-parameters for AMOTA (shown in the table with ``$\Psi$,'' details are provided in Sec.~\ref{sup:sec:tracking_by_detection} (appendix). The results clearly demonstrate the advantages of our end-to-end approach compared to more traditional, modular frameworks, with improvements being especially significant on the ID-Switch metric.

\input{tables/tracking_by_detection.tex}

\input{tables/e2e_prediction.tex}

\mypar{Analysis on Prediction from Query Features.} 
While our paper focuses on multi-object tracking, we additionally provide an analysis of prediction performance.
We show that predicting end-to-end from object features is advantageous over predicting from low-level object states, such as center positions. Specifically, we train two motion prediction baselines, VectorNet~\cite{gao2020vectornet} and LSTM~\cite{chang2019argoverse}, using the true-positive tracks from \name\ following previous studies~\cite{luo2018fast}, and report their results in the top rows of Tab.~\ref{tab:e2e_prediction}. As our method does not use HD-Maps, for a fair comparison, we exclude the parts of motion prediction algorithms that handle HD-Maps in these experiments. In addition, we report another baseline which uses the velocities predicted by our model's decoder for trajectory prediction, assuming a constant velocity motion model (third row in Tab.~\ref{tab:e2e_prediction}). The evaluation metrics are ``average displacement error'' (ADE) and ``final displacement error'' (FDE), which are better with lower values. More details are in Sec.~\ref{sup:sec:e2e_motion_prediction} (appendix).

Tab.~\ref{tab:e2e_prediction} compares the performance between the end-to-end \name\ and the baselines described above on the validation split of nuScenes. With lower ADE and FDE, \name\ has better trajectory quality. Our conclusions agree with previous studies~\cite{gu2022vip3d, weng2022mtp, luo2018fast}. More specifically, LSTM is a shallow model and unable to capture meaningful dynamics from noisy tracks; the stronger VectorNet model can perform better than the other baselines, but it is still worse than forecasting trajectories in an end-to-end framework, as proposed in our method.

\subsection{Qualitative Results}

We visualize the 3D MOT results in Fig.~\ref{fig:vis_track} by projecting 3D bounding boxes onto images. The colors of bounding boxes are randomly selected from a pool of seven colors according to their IDs, so that each object has a consistent color over time. 

In the top row, we provide an overall visualization of multi-camera 3D MOT, focusing on front-left and front cameras. As clearly shown, \name\ tracks objects coherently, especially for the pedestrians and vehicles shown on two separate cameras. In the bottom two rows of Fig.~\ref{fig:vis_track}, we illustrate two examples of addressing occlusions. For both large (bus) and small (pedestrian) objects, our method propagates their positions during the occluded frames and successfully re-associates them on de-occlusion frames even on a different camera. We highlight that \emph{this is achieved without an explicit re-identification module}. 

%% file: tables/sota_comparison.tex
\begin{table}[t]
\centering
\resizebox{1.0\linewidth}{!}{
\begin{tabular}{l@{\hspace{2mm}}|c@{\hspace{1.2mm}} c@{\hspace{1.2mm}} c@{\hspace{1.2mm}} c@{\hspace{1.2mm}} c@{\hspace{1.2mm}}}
\specialrule{1pt}{0pt}{1pt}

 & AMOTA $\uparrow$ & AMOTP $\downarrow$ & RECALL $\uparrow$ & MOTA $\uparrow$ & IDS $\downarrow$ \\

\midrule
\textit{Validation Split} \\
\midrule

DEFT~\cite{chaabane2021deft} & 0.201 & N/A & N/A &  0.171 & N/A \\
QD3DT~\cite{hu2022monocular} & 0.242 & 1.518 & 39.9\% &  0.218 & 5646 \\
MUTR3D~\cite{mutr3d}  & 0.294 & 1.498 & 42.7\%  & 0.267  & 3822 \\
TripletTrack~\cite{marinello2022triplettrack} & 0.285 & 1.485 & N/A & N/A & N/A \\
CC-3DT$^*$~\cite{cc3dt} & 0.429 & 1.257 & 53.4\% & 0.385 & 2219 \\
\midrule
\name-S (Ours) & 0.408 & 1.343 & 50.7\% & 0.376 & \textbf{166} \\
\name-F (Ours) & \textbf{0.479} & \textbf{1.227} & \textbf{59.0\%} & \textbf{0.435} & 181 \\
\midrule
\textit{Test Split} \\
\midrule
CenterTrack~\cite{centertrack} & 0.046 & 1.543 & 23.3\% & 0.043 & 3807 \\
PermaTrack~\cite{tokmakov2021learning} & 0.066 & 1.491 & 18.9\% & 0.060 & 3598 \\
DEFT~\cite{chaabane2021deft} & 0.177 & 1.564 & 33.8\% & 0.156 &  6901  \\
QD3DT~\cite{hu2022monocular} & 0.217 & 1.550 & 37.5\% &  0.198 & 6856 \\
MUTR3D~\cite{mutr3d} & 0.270 & 1.494 & 41.1\%  & 0.245  & 6018 \\
TripletTrack~\cite{marinello2022triplettrack} & 0.268 & 1.504 & 40.0\% & 0.245 & 1144 \\
CC-3DT$^*$~\cite{cc3dt} & 0.410 & 1.274 & \textbf{53.8\%} & 0.357 & 3334 \\
\midrule
\name-F (Ours) & \textbf{0.434} & \textbf{1.252} & \textbf{53.8\%} & \textbf{0.378} & \textbf{249} \\
\bottomrule
\end{tabular}
}
\vspace{-2mm}
\caption{Comparison with state-of-the-art camera-based 3D MOT algorithms on nuScenes~\cite{nuscenes}. ``\textbf{S}'' and ``\textbf{F}'' denotes our model trained with small-resolution and full-resolution setting, respectively (clarified in Sec.~\ref{sec:implementation_details}). Our approach has a significant advantage on both AMOTA and ID-Switch (full-resolution), where ID-Switch is almost 90\% less and \textit{an order of magnitude smaller} compared to other methods.
(*) indicates concurrent works.
\vspace{-4mm}
}
\label{tab:sota}
\end{table}

%% file: tables/full_ablations.tex
\begin{table}[tb]
\resizebox{0.99\linewidth}{!}{
\begin{tabular}{c@{\hspace{1.2mm}}|c@{\hspace{0.8mm}}c@{\hspace{0.8mm}}|c@{\hspace{0.8mm}}c@{\hspace{0.8mm}}|c@{\hspace{1.2mm}}c@{\hspace{1.2mm}}c@{\hspace{1.2mm}}}
\hline
\multicolumn{1}{c|}{\multirow{2}{*}{Index}} & \multicolumn{2}{c|}{Past} & \multicolumn{2}{c|}{Future} & \multicolumn{1}{c}{\multirow{2}{*}{AMOTA$\uparrow$}} & \multicolumn{1}{c}{\multirow{2}{*}{AMOTP$\downarrow$}} & \multicolumn{1}{c}{\multirow{2}{*}{IDS$\downarrow$}} \\
\multicolumn{1}{c|}{} & \multicolumn{1}{c}{QR} & \multicolumn{1}{l|}{TR} & \multicolumn{1}{c}{Pred} & \multicolumn{1}{c|}{Ext} & \multicolumn{1}{c}{} & \multicolumn{1}{c}{} & \multicolumn{1}{c}{} \\ \midrule
1 &  &  &  &  & 0.368 & 1.421 & 507 \\
2 & \cmark & & & & 0.378 & 1.414 & 453 \\
3 & \cmark & \cmark & & & 0.380 & 1.408 & 400 \\
4 & & & \cmark & & 0.374 & 1.402 & 469 \\
5 & & & \cmark & \cmark & 0.391 & 1.360 & \textbf{155} \\
6 & \cmark & \cmark & \cmark & \cmark & \textbf{0.408} & \textbf{1.343} & 166 \\
\bottomrule
\end{tabular}
}
\caption{\textbf{Ablation of \name\ Modules.} For past reasoning, ``QR'' and ``TR'' denote ``query refinement'' and ``track refinement'' in Sec.~\ref{sec:past}. For future reasoning, ``Pred'' and ``Ext'' denote ``motion prediction'' and ``track extension'' in Sec.~\ref{sec:future}. Past and future reasoning improve 3D MOT independently, and \name\ achieves top results by combining them in an end-to-end framework.}
\vspace{-4mm}
\label{tab:ablation}
\end{table}

%% file: tables/prediction_length.tex
\begin{table}[t]
\centering
\resizebox{0.83\linewidth}{!}{
\begin{tabular}{c| c|  c c c }
\specialrule{1pt}{0pt}{1pt}

   Length & Extention & AMOTA $\uparrow$ & AMOTP $\downarrow$ & IDS $\downarrow$ \\

\midrule
2.0s & \xmark  & 0.392 & 1.376 & 604 \\
2.0s & \cmark & 0.402 & 1.342 & 217 \\
\midrule
3.0s & \xmark & 0.392 & 1.372 & 540   \\
3.0s & \cmark & 0.402 & \textbf{1.340} & 208  \\
\midrule
4.0s & \xmark  & 0.391 & 1.387  & 471 \\
4.0s & \cmark & \textbf{0.408} & 1.343 & \textbf{166}  \\
\bottomrule
\end{tabular}
}
\caption{\textbf{Length of motion prediction}. ``Extension'' means using ``track extension.'' We train three models with the prediction horizon of 2.0s, 3.0s, and 4.0s. According to AMOTA and IDS, learning a longer prediction benefits tracking.
}
\vspace{-2mm}
\label{tab:prediction_length}
\end{table}

%% file: tables/tracking_by_detection.tex
\begin{table}[t]
\centering
\resizebox{0.83\linewidth}{!}{
\begin{tabular}{l| c c c }
\specialrule{1pt}{0pt}{1pt}

 & AMOTA $\uparrow$ & AMOTP $\downarrow$ & IDS $\downarrow$ \\
\midrule
AB3DMOT~\cite{ab3dmot} & 0.292 & 1.333 & 2419 \\
AB3DMOT~\cite{ab3dmot}$^\Psi$ & 0.329 & 1.388 & 2677 \\
\midrule
CenterPoint~\cite{yin2021center} & 0.233 & \textbf{1.270} & 2715 \\
CenterPoint~\cite{yin2021center}$^\Psi$ & 0.383 & 1.329 & 3082 \\
\midrule
SimpleTrack~\cite{simpletrack} & 0.320 & 1.295 & 1606 \\
SimpleTrack~\cite{simpletrack}$^\Psi$ & 0.402 & 1.324 & 2053 \\
\midrule
\name\ (Ours) & \textbf{0.408} & 1.343 & \textbf{166} \\

\bottomrule
\end{tabular}}
\caption{\textbf{Comparison with ``tracking by detection.''} We apply strong baselines in 3D MOT to PETR~\cite{petr}: AB3DMOT~\cite{ab3dmot}, CenterPoint~\cite{yin2021center}, and SimpleTrack~\cite{simpletrack}. ``$\Psi$'' means that we tune the hyper-parameters of these methods to fit PETR detections, rather than using their original configuration. Our end-to-end approach has significant advantages. }
\vspace{-2mm}
\label{tab:tracking_by_detection}
\end{table}

%% file: tables/e2e_prediction.tex
\begin{table}[tb]
    \centering
\scalebox{0.80}{
    \begin{tabular}{l|cc}
    \toprule
         Method & ADE $\downarrow$ (@4.0s) & FDE $\downarrow$ (@4.0s) \\
    \midrule
         LSTM~\cite{chang2019argoverse} & 2.32 & 2.87 \\
         VectorNet~\cite{gao2020vectornet} & 2.01 & 2.48 \\
    \midrule
    Velocity & 2.10 & 2.64 \\
         \name\ (Ours) & \textbf{1.88} & \textbf{2.38} \\
    \bottomrule
    \end{tabular}
}
    \caption{\textbf{Motion prediction from features or abstract states.} We build a motion prediction benchmark from the true-positive tracks of \name\ on the nuScenes validation split, and then train LSTM~\cite{chang2019argoverse} and VectorNet~\cite{gao2020vectornet} from the 3D positions of tracks. The ``Velocity'' row is the result under the assumption of a constant velocity motion model. The results indicate that predicting from features provides richer information for better trajectory quality.}
    \vspace{-2mm}
    \label{tab:e2e_prediction}
\end{table}

%% file: sections/5-conclusions.tex
\section{Conclusions}
This paper proposes a query-based end-to-end method for multi-camera 3D MOT that enhances spatio-temporal coherence. By past reasoning, our framework enhances the query features and track quality with historical information. By future reasoning, the predicted trajectories better propagate the queries across adjacent frames and occluded long-term periods. We also demonstrate that joint past and future reasoning further strengthens the tracker's ability. Extensive evaluation of the large-scale nuScenes dataset demonstrates that our method is effective in providing coherent tracks.

\small{\mypar{Acknowledgement.} This work was supported in part by Toyota Research Institute, NSF Grant 2106825, NIFA Award 2020-67021-32799, and the NCSA Fellows program.}

%% file: supp_sections/implementation.tex
\section{Implementation Details}
\label{sec:supp_implementation_details}
\subsection{Model Architecture}
\label{sec:supp_model_arch}
We explain the design choices of \name\ in the sections below. To provide a high-level view of the model, we enclose the config file in mmdetection3d~\cite{mmdet3d2020} format in this supplementary material.

\mypar{Backbone.} We use VoVNetV2~\cite{lee2019energy} as backbone. For the feature pyramid~\cite{lin2017feature}, the C5 feature (output of the 5-th stage) is upsampled and fused with C4 feature (output of the 4-th stage). To save GPU memory during training, we adopt the checkpointing trick~\cite{chen2016training} by default.

\mypar{Detection head.} We follow the design of PETR~\cite{petr} by setting the region to $[-51.2m, 51.2m]$ on the XY-axis and $[-5m, 3m]$ on the Z-axis. The centers of bounding boxes are normalized to $[0, 1]$, respectively. The detection head composes of 6 transformer decoder layers~\cite{vaswani2017attention} and 2 MLP heads for bounding box regression and classification. Each transformer decoder layer has an embedding dimension of 256 and a feedforward dimension of 2048. The dropout probability is 0.1. The MLP heads are both two-layer MLPs. The bounding box regression head predicts the centers $(x, y, z)$, sizes $(l, w, h)$, orientation, and velocities of objects, and the classification head returns the logits for every category.

\mypar{\name\ configurations.} Our \name\ uses a fixed number of 500 detection queries per frame. As for track queries, the training and inference procedures are different. Training protocol adds queries into the set of track queries once they become a positive match to the ground truth ($<$2.0m), and a tracking query is kept associated with the ground truth of the same object. The inference protocol initializes the track queries if the corresponding confidence score is larger than 0.4. During the inference time, our model output at most 300 objects (same as~\cite{detr3d, petr, mutr3d}) and set a minimum score threshold of 0.2.

\mypar{Past reasoning.} The length of the query queue  is 1.5 seconds ($\tau_h=3$ frames) because each of our training samples has 3 frames. However, we emphasize that \name\ is able to aggregate historical information from the entire video during the \textit{inference} time because the past reasoning module is recurrent. Concretely, the queries at frame $t$ attend to frames $[t-2, t]$, and the queries at frame $t-2$, in turn, attend to frames $[t-4, t-2]$, and so on. Thus, the queries at frame $t$ have access to information from all the previous frames.

For cross-frame and cross-frame attention, we employ two transformer decoder layers for them each. Every transformer decoder layer has an embedding dimension of 256, and a feedforward dimension of 2048. Same as the transformer layers in the detection head, these two layers also have a dropout probability of 0.1. The track refinement module uses two separate 2-layer MLP heads for regression and classification. 

\mypar{Future reasoning.} The future reasoning module predicts the movements to future 4.0 seconds ($\tau_f=8$ frames). It first uses the same cross-frame attention to generate the future features, then applies a 2-layer MLP to translate features into the movements on the XY plane. For track extension, we pick the extension length that maximizes the AMOTA on the validation split, which are $2.0s$ (4 frames) for the small resolution model and $2.5s$ (5 frames) for the full resolution model.

\mypar{Loss weights.} In Sec.~\ref{sec:loss} of the main paper, we use coefficients to balance the loss terms. The bounding box regression loss is $\lambda^D_{box}=0.25$, and the classification focal loss is $\lambda^D_{cls}=2.0$, which are the same as~\cite{petr}. For the track refinement part, we adopt the same loss weights for bounding box regression and classification, respectively: $\lambda^R_{box}=0.25$ and $\lambda^R_{cls}=2.0$. The weight for motion prediction in future reasoning is $\lambda_{f} = 0.5$.

\input{supp_sections/algo_prop}

\subsection{Algorithm for ``Track Extension''}
\label{sup:sec:algo_track_extension}

We clarify the detailed steps for the track extension algorithms described in Sec.~\ref{sec:future} (main paper). For best clarity, we rigorously describe query propagation in Algorithm~\ref{algo:propagation} first and then introduce track extension in Algorithm~\ref{algo:track_extension}.

\input{supp_sections/algo_track_extension}

As in Algorithm~\ref{algo:propagation}, the propagated center positions $\hat{\textbf{C}}_{t-1}$ come from adding current states $\textbf{C}_{t-1}$ with motion predictions $\textbf{M}^{t-1}_{t-1:t}$. The propagation of queries $\hat{\textbf{Q}}_{t-1}$ directly re-use the latest results $\textbf{Q}_{t-1}$. The propagation of motion predictions can be simple padding functions, as our track extension does not exceed the length of motion prediction.

Then we illustrate the algorithm after adding track extension in Algorithm~\ref{algo:track_extension}. The major difference lies in addressing the low-confidence objects (lines 3-14), which might be noisy. Instead of always updating according to the latest results, track extension relies more on the results coming from previous frames for better fidelity (lines 5-8). Notably, if an object does not have confident results in $\tau_e$ continuous frames, we follow the common practice of terminating this track (lines 10-13). Please note that we set $\tau_e$ smaller than the prediction horizon $\tau_f=8$ frames (4.0s) by default. As in Tab.~\ref{tab:ablation}, track extension effectively improves both AMOTA and IDS.

\subsection{Model Training}

Every training sample composes of three adjacent frames. \name\ is trained with AdamW optimizer~\cite{kingma2014adam, loshchilov2017decoupled} with an weight decay of 0.01. The learning rate starts from $2.0\times 10^{-4}$ and is scheduled according to cosine annealing~\cite{loshchilov2016sgdr}. The above process is identical to image-based 3D detection methods~\cite{detr3d, petr}. However, as image data augmentation could break the motion models of objects, we disable the data augmentation during the training of the tracker. The total training epochs follow the settings discussed in Sec.~\ref{sec:implementation_details} (main paper). The full resolution setting takes 3 days on 8$\times$A100 GPUs, and the small resolution setting takes 1 day on 8$\times$A100 GPUs.

\subsection{``Tracking by Detection'' Experiments}
\label{sup:sec:tracking_by_detection}

The experiments of ``tracking by detection'' baselines in Sec.~\ref{sec:ablation} (main paper) tune the detection score threshold according to AMOTA on the validation split. Eventually, we set a minimum score threshold of 0.2 for output bounding boxes, which is also the same as \name. According to Tab.~\ref{tab:tracking_by_detection} (main paper), our hyper-parameter tuning significantly improves the performance of ``tracking by detection'' baselines and enables a fair comparison. The reason is that AB3DMOT~\cite{ab3dmot}, CenterPoint~\cite{yin2021center}, and SimpleTrack~\cite{simpletrack} designed their trackers for LiDAR-based 3D detection and used low (\textit{e.g.}, 0.01 in SimpleTrack) or no score thresholds. However, image-based detection contains more false positives and requires stronger filtering.

\subsection{Motion Prediction from Abstract Object States Settings}
\label{sup:sec:e2e_motion_prediction}
This section explains the details of the ``Prediction from Query Features'' in Sec.~\ref{sec:ablation} (main paper). 

\mypar{Dataset construction.} We construct the motion prediction dataset by recalling the true positive tracks from the output of \name. Specifically, we perform Hungarian matching between the predicted bounding boxes and ground truth, and the positive matches are determined by less than 2.0m from the ground truth. We use the training/validation split of nuScenes~\cite{nuscenes} dataset and remove the frames without positive match. Eventually, we have 27,960 and 5,879 frames, and 422,167 and 69,804 tracks in the training and validation set for motion prediction.

\mypar{Model training and inference.} Our model architectures are the same as the LSTM model from the Argoverse~\cite{chang2019argoverse} and VectorNet~\cite{gao2020vectornet}. We provide 2.0s of history and require the model to predict up to 4.0s, which is the same as our \name, for a fair comparison. As every frame contains multiple tracks, our implementation normalizes the coordinates with respect to the positions of ego-vehicle, following~\cite{ettinger2021large}. The inputs to the models include the bounding box centers and velocities on the XY plane. For missing observations, we pad the input to the full length with the closest observation and indicate padding with a 0-1 mask. We train the models for 24 epochs on our custom motion prediction dataset with AdamW~\cite{kingma2014adam, loshchilov2017decoupled} optimizer and an initial learning rate of $1.0 \times 10^{-3}$. The learning rate drops by $0.1$ on the 16-th and 20-th epochs.

%% file: supp_sections/algo_prop.tex
\begin{algorithm}[t]
\caption{Algorithm for ``Query Propagation.''}
\begin{algorithmic}[1]
\label{algo:propagation}
\renewcommand{\algorithmicrequire}{\textbf{Input:}}
\renewcommand{\algorithmicensure}{\textbf{Output:}}
\REQUIRE \ \\$\textbf{Q}_{t-1}$, $\textbf{C}_{t-1}$, $\textbf{M}^{t-1}_{t-1:t+\tau_f-1}$: queries, center positions, and motion predictions of objects from frame $t-1$; \\
$N$: number of queries.
\ENSURE \ \\ $\hat{\textbf{Q}}_{t-1}$, $\hat{\textbf{C}}_{t-1}$, $\hat{\textbf{M}}^{t-1}_{t:t+\tau_f}$: propagated queries, center positions, and motion predictions of objects from frame $t-1$ to frame $t$.
\\ \hrulefill
\FOR{$i=$ 0 to $N-1$ }
\STATE $\hat{\textbf{C}}^i_{t-1} \xleftarrow{} \textbf{C}^{i}_{t-1} + \textbf{M}^{t-1, i}_{t-1:t}$
\STATE $\hat{\textbf{Q}}^i_{t-1} \xleftarrow{} \textbf{Q}^{i}_{t-1}$
\STATE $\hat{\textbf{M}}^{t-1}_{t:t+\tau_f}\xleftarrow{}\mathtt{Padding}(\textbf{M}^{t-1}_{t:t+\tau_f-1})$
\ENDFOR
\RETURN $\hat{\textbf{Q}}_{t-1}$, $\hat{\textbf{C}}_{t-1}$, $\hat{\textbf{M}}^{t-1}_{t:t+\tau_f}$
\end{algorithmic}
\end{algorithm}

%% file: supp_sections/algo_track_extension.tex
\begin{algorithm}[t]
\caption{Algorithm for ``Track Extension''.}
\begin{algorithmic}[1]
\label{algo:track_extension}
\renewcommand{\algorithmicrequire}{\textbf{Input:}}
\renewcommand{\algorithmicensure}{\textbf{Output:}}
\REQUIRE \ \\$\hat{\textbf{Q}}_{t-1}$, $\hat{\textbf{C}}_{t-1}$, $\hat{\textbf{M}}^{t-1}_{t:t+\tau_f}$: propagated queries, center positions, and motion predictions from frame $t-1$; \\
$\textbf{F}_t$: image features on frame $t$; \\ 
$\textbf{L}_t$: how many frames have the queries been extended continuously; \\ 
$\tau_{e}$: maximum frames for extension; \\
$N$: number of queries.
\ENSURE \ \\ $\hat{\textbf{Q}}_{t}$, $\hat{\textbf{C}}_{t}$, $\hat{\textbf{M}}^{t}_{t+1:t+\tau_f+1}$: propagated queries, center positions, and motion predictions of tracked objects from frame $t$ to frame $t+1$.
\\ \hrulefill
\STATE $\textbf{Q}^R_{t}, \textbf{C}^R_t, \textbf{M}_{t:t+\tau_f}^{t} \xleftarrow{} \text{PF-Track}(\textbf{F}_t, \textbf{Q}_t, \textbf{C}_t)$
\FOR{$i=$ 0 to $N-1$ }
\IF{confidence score $S^{R, i}_t$ is below a threshold}
\IF{$i$ is an active object $L_t^i < \tau_e$}
\STATE Then use previous information by Line 6-8:
\STATE $\hat{\textbf{C}}^i_t \xleftarrow{} \hat{\textbf{C}}^{i}_{t-1} + \hat{\textbf{M}}^{t-1, i}_{t:t+1}$
\STATE $\hat{\textbf{Q}}^i_t\xleftarrow{} \hat{\textbf{Q}}^i_{t-1}$
\STATE $\hat{\textbf{M}}^{t}_{t+1:t+\tau_f+1}\xleftarrow{}\mathtt{Padding}(\hat{\textbf{M}}^{t-1}_{t+1:t+\tau_f})$
\STATE Record the extension: $L_t^i \xleftarrow{} L_t^i + 1$
\ELSE
\STATE Terminate the track of the $i$-th object.
\STATE Remove it from memory.
\ENDIF

\ELSE
\STATE Propagate normally as Line 15-18
\STATE $\hat{\textbf{C}}^i_t \xleftarrow{} \textbf{C}^{R, i}_t + \textbf{M}^{t, i}_{t:t+1}$
\STATE $\hat{\textbf{Q}}^i_t\xleftarrow{} \textbf{Q}^{R, i}_{t}$
\STATE $\hat{\textbf{M}}^{t}_{t+1:t+\tau_f+1}\xleftarrow{}\mathtt{Padding}(\textbf{M}^{t}_{t+1:t+\tau_f})$
\STATE Zero the continuous extension: $L_t^i \xleftarrow{} 0$
\ENDIF
\ENDFOR
\RETURN $\hat{\textbf{Q}}_{t}$, $\hat{\textbf{C}}_{t}$, $\hat{\textbf{M}}^{t}_{t+1:t+\tau_f+1}$
\end{algorithmic}
\end{algorithm}

%% file: supp_sections/more_ablation.tex
\section{Supplemental Ablation Studies}
\label{sec:supp_ablation}

\subsection{Cross-frame and Cross-object Attention in Past Reasoning.}

We analyze the effect of cross-frame and cross-object attention for ``Query Refinement'' in past reasoning (Sec.~\ref{sec:past} of main paper). As in Tab.~\ref{tab:cross_frame_cross_object}, if either of them is removed from the final \name, the performance decreases, especially in AMOTA. Although \name\ has a slightly larger ID-Switch, we argue that it is due to a higher AMOTA. Therefore, both cross-frame and cross-object attentions are useful for multi-camera 3D MOT.

\input{tables/cross_frame_cross_object.tex}

\input{tables/velo_prop.tex}

\subsection{Trajectories or Velocity for Query Propagation}

To verify the necessity of learning a trajectory prediction for query propagation, we experiment with using the velocities for propagation in Tab.~\ref{tab:velo_prop}. Specifically, we train a model identical to \name\ except using velocities to transform the positions of queries across frames. As in Tab.~\ref{tab:velo_prop}, using learned trajectories has a better performance compared with the variant of using velocities. Therefore, we conclude that learning long-term trajectory prediction is necessary for robust and accurate 3D MOT.

\subsection{Weights of Motion Prediction in Future Reasoning}

We analyze the tracking performance with respect to the loss weights of motion prediction. The discovery is the sensitivity of tracking performance to the weight of the loss for motion prediction $\lambda_f$ (Sec.~\ref{sec:loss} in main paper). In Tab.~\ref{tab:prediction_weights}, we vary the weight for motion prediction with $\lambda_f=[0.25, 0.50, 1.00]$, and they could cause variation in the tracking performance. This requires the attention of future works or better multi-task learning strategies

\input{tables/prediction_weights.tex}

\subsection{Category-level Analysis}

\begin{figure}[tb]
    \centering
    \includegraphics[width=0.7\linewidth]{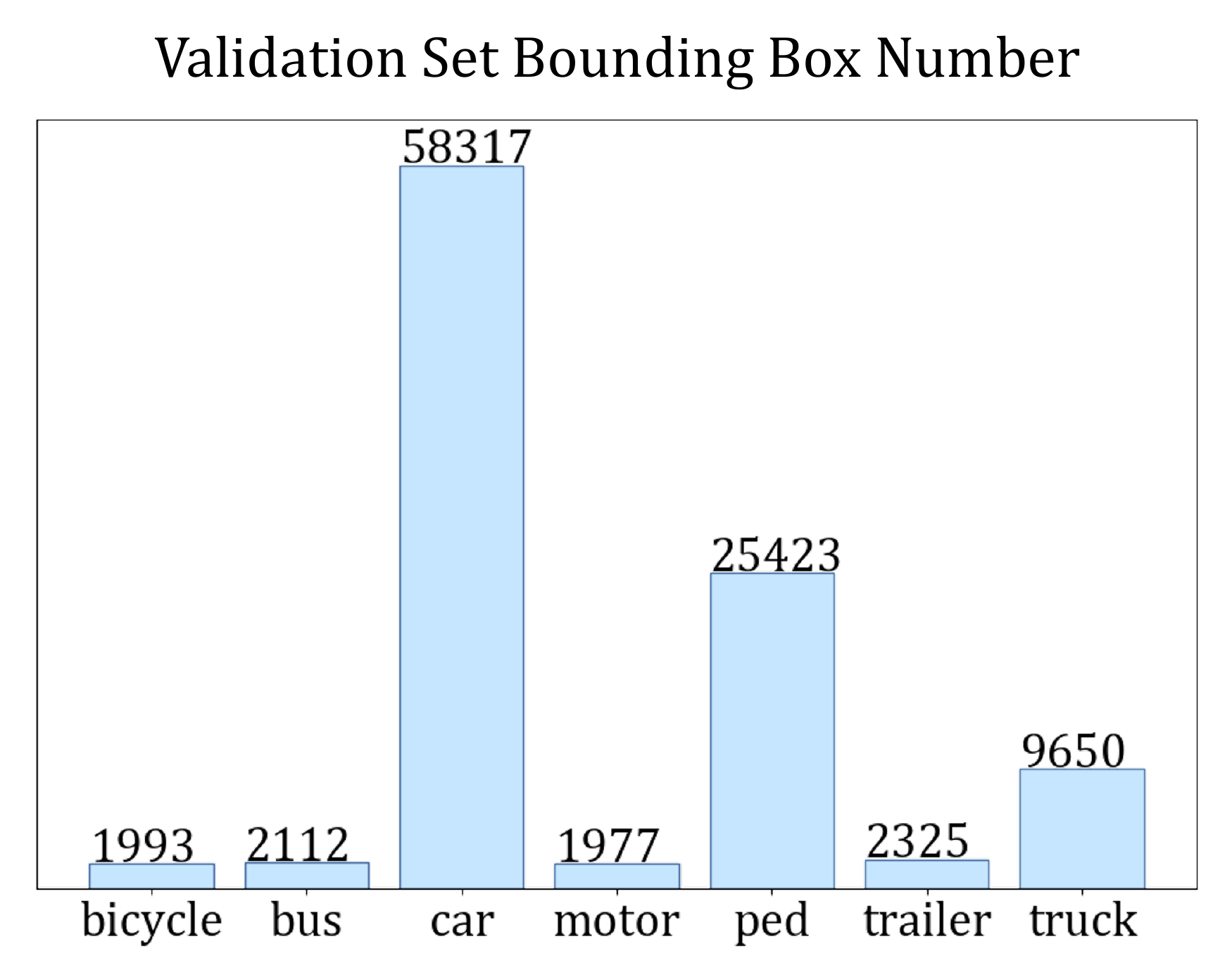}
    \vspace{-3mm}
    \caption{Distribution of per-category instance numbers in nuScenes validation set. ``motor'' denotes ``motorcycle'' and ``ped'' denotes ``pedestrian.'' As clearly demonstrated, the data distribution is imbalanced on nuScenes.}
    \vspace{-4mm}
    \label{fig:cat_distribution}
\end{figure}

We compare the category-level performance before and after using our past and future reasoning. To provide the context of nuScenes dataset~\cite{nuscenes}, we start by visualizing the data distribution across categories in Fig.~\ref{fig:cat_distribution}. As clearly illustrated, nuScenes exhibits an imbalanced category distribution, and the types of ``car'' and ``pedestrian'' take up most of the objects.

Then in Tab.~\ref{tab:category}, we compare the category-level performance of the three major types: car, pedestrian, and truck. Compared to the baseline of not using any past or future reasoning (same baseline as Tab.~\ref{tab:sota} of the main paper), our proposed method significantly improves upon the baseline over individual categories.

\input{tables/category_analysis.tex}

%% file: tables/cross_frame_cross_object.tex
\begin{table}[tb]
\centering
\resizebox{1.0\linewidth}{!}{
\begin{tabular}{c| c| c c c }
\specialrule{1pt}{0pt}{1pt}

  Model & Extension & AMOTA $\uparrow$ & AMOTP $\downarrow$ & IDS $\downarrow$ \\
   \midrule
 \multirow{2}{*}{\name} &  & 0.391 & 1.387 & 471 \\
 & \cmark & \textbf{0.408} & \textbf{1.343} & 166 \\
 \midrule
 \multirow{2}{*}{wo/ Cross-frame} &  & 0.385 & 1.402 & 410 \\
 & \cmark & 0.395 & 1.373 & 171 \\
 \midrule
 \multirow{2}{*}{wo/ Cross-object} &  & 0.383 & 1.390 & 481 \\
 & \cmark & 0.399 & 1.352 & \textbf{155} \\
\bottomrule
\end{tabular}
}
\vspace{-3mm}
\caption{\textbf{Cross-frame and Cross-object attention}. ``Extension'' denotes using track extension or not. Removing either cross-frame or cross-object from the ``Query Refinement'' module of past reasoning (Sec.~\ref{sec:past} of main paper) negatively impacts the model performance.
}
\vspace{-4mm}
\label{tab:cross_frame_cross_object}
\end{table}

%% file: tables/velo_prop.tex
\begin{table}[tb]
\centering
\resizebox{0.9\linewidth}{!}{
\begin{tabular}{c| c| c c c }
\specialrule{1pt}{0pt}{1pt}

  Model & Extension & AMOTA $\uparrow$ & AMOTP $\downarrow$ & IDS $\downarrow$ \\
   \midrule
 \multirow{2}{*}{\name} &  & 0.391 & 1.387 & 471 \\
 & \cmark & \textbf{0.408} & \textbf{1.343} & \textbf{166} \\
 \midrule
 \multirow{2}{*}{Velo Prop} &  & 0.374 & 1.389 & 478 \\
 & \cmark & 0.382 & 1.359 & 192 \\
\bottomrule
\end{tabular}
}
\vspace{-3mm}
\caption{\textbf{Velocity for Query Propagation}. ``Extension'' means using track extension. Using velocities instead of learned trajectories for query propagation in future reasoning (Sec.~\ref{sec:future} of main paper) negatively impacts the model performance.
}
\vspace{-4mm}
\label{tab:velo_prop}
\end{table}

%% file: tables/prediction_weights.tex
\begin{table}[tb]
\centering
\resizebox{0.8\linewidth}{!}{
\begin{tabular}{c| c| c c c }
\specialrule{1pt}{0pt}{1pt}

  $\lambda_f$ & Extension & AMOTA $\uparrow$ & AMOTP $\downarrow$ & IDS $\downarrow$ \\
   \midrule
 \multirow{2}{*}{0.25} &  & 0.389 & 1.382 & 539 \\
 & \cmark & 0.396 & 1.353 & 185 \\
 \midrule
 \multirow{2}{*}{0.50} &  & 0.391 & 1.387 & 471 \\
 & \cmark & \textbf{0.408} & \textbf{1.343} & \textbf{166} \\
 \midrule
 \multirow{2}{*}{1.00} & \ & 0.377 & 1.395 & 489 \\
 & \cmark & 0.397 & 1.346 & 183 \\
\bottomrule
\end{tabular}
}
\vspace{-3mm}
\caption{\textbf{Loss weights of motion prediction}. ``$\lambda_f$'' denotes the loss weight $\lambda_f$ for motion prediction (Sec.~\ref{sec:loss} in the main paper), ``Extension'' denotes using track extension or not. This table demonstrates the sensitivity of motion prediction and calls for the attention of future work in better multi-task learning strategies.
}
\vspace{-4mm}
\label{tab:prediction_weights}
\end{table}

%% file: tables/category_analysis.tex
\begin{table}[t]
\centering
\resizebox{0.9\linewidth}{!}{
\begin{tabular}{c| c| c c c | c}
\specialrule{1pt}{0pt}{1pt}

  Metrics & \name & Car & Pedestrian  & Truck & All \\
   \midrule
 \multirow{2}{*}{AMOTA $\uparrow$} &  & 0.562 & 0.372 & 0.374 & 0.358 \\
 & \cmark &  \textbf{0.579} & \textbf{0.415} & \textbf{0.403} & \textbf{0.408} \\
 \midrule
 \multirow{2}{*}{AMOTP $\downarrow$} &  & 1.058 & 1.431 & 1.341 & 1.419 \\
 & \cmark & \textbf{1.021} & \textbf{1.362} & \textbf{1.288} & \textbf{1.343} \\
 \midrule
 \multirow{2}{*}{IDS $\downarrow$} & & 368 & 154 & 25 & 507 \\
 & \cmark & \textbf{67} & \textbf{83} & \textbf{7} & \textbf{166} \\
\bottomrule
\end{tabular}
}
\vspace{-2mm}
\caption{\textbf{Per-category tracking metrics analysis}. ``All'' denotes the averaged metric numbers over all seven categories. On the three major categories on nuScenes, our \name\ achieves significant and steady improvement over the baseline that does not involve past or future reasoning.
}
\vspace{-4mm}
\label{tab:category}
\end{table}

%% file: supp_sections/detr3d.tex
\begin{figure*}[t]
    \centering
    \includegraphics[width=1.0\textwidth]{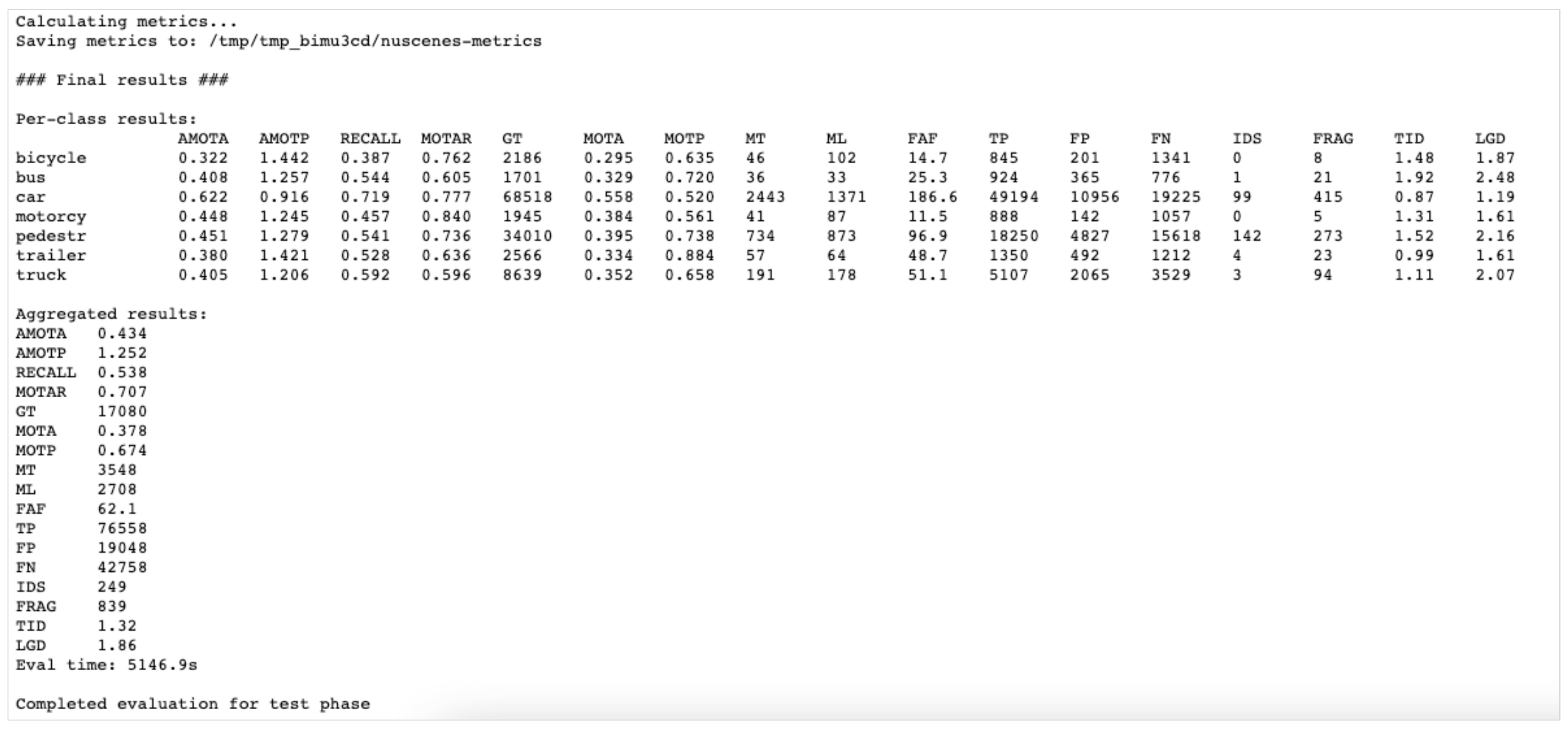}
    \caption{\textbf{Screenshot of test set result.} This is in supplementary for our results in Tab.~\textcolor{blue}{1} (main paper).}
    \label{fig:screenshot}
\end{figure*}

\subsection{PF-Track with DETR3D}
\label{sec:supp_detr3d}

\input{tables/detr3d.tex}

We provide supplemental analysis of applying \name\ to a different 3D detector: DETR3D~\cite{detr3d}.

\mypar{Experiment setup.} Following DETR3D~\cite{detr3d}, we use VoVNetV2~\cite{lee2019energy} as the backbone and fuse the features from C2-C5 with FPN~\cite{lin2017feature} as the image features. As DETR3D only has a full-resolution setting, we first train the backbone with single-frame detection for 24 Epochs, then fix the backbone and train the whole tracker with 3-frame tracking for another 24 Epochs. A cycle of training takes 3 days on 8 $\times$ A100 GPUs.

\mypar{Efficacy of \name.} We analyze the performance without or with the joint past and future reasoning in \name\ in Tab.~\ref{tab:detr3d}. We include the results for top-3 major categories and the average of all categories. Our past and future reasoning also significantly improves the AMOTA and decreases ID-Switches. 
This result indicates the generalizability of \name\ on other query-based detectors. 


%% file: tables/detr3d.tex
\begin{table}[tb]
\centering
\resizebox{0.9\linewidth}{!}{
\begin{tabular}{c| c| c c c | c}
\specialrule{1pt}{0pt}{1pt}

  Metrics & \name & Car & Pedestrian  & Truck & All \\
   \midrule
 \multirow{2}{*}{AMOTA $\uparrow$} &  & 0.583 & 0.370 & 0.305 & 0.344 \\
 & \cmark &  \textbf{0.600} & \textbf{0.431} & \textbf{0.310} & \textbf{0.362} \\
 \midrule
 \multirow{2}{*}{AMOTP $\downarrow$} &  & 1.038 & 1.436 & 1.399 & 1.419 \\
 & \cmark & \textbf{1.001} & \textbf{1.338} & \textbf{1.366} & \textbf{1.363} \\
 \midrule
 \multirow{2}{*}{IDS $\downarrow$} &  & 394 & 222 & 32 & 680 \\
 & \cmark & \textbf{172} & \textbf{116} & \textbf{5} & \textbf{300} \\
\bottomrule
\end{tabular}
}
\vspace{-2mm}
\caption{\textbf{\name\ for DETR3D detection head}. ``All'' denotes the averaged metric numbers over all seven categories, ``\name'' denotes using the past and future reasoning from \name. On the three major categories and average metrics on nuScenes, \name\ is able to enhance the tracking performance.
}
\vspace{-4mm}
\label{tab:detr3d}
\end{table}

%% file: supp_sections/miscellaneous.tex
\section{Limitations and Future Works}
\vspace{-2mm}

\mypar{Processing HD-Map for End-to-end Forecasting.}
The main focus of \name\ is 3D MOT, and it is not a full-fledged motion prediction pipeline because of not consider HD-Maps. However, with the improved track quality, our past and future reasoning could be beneficial to downstream motion prediction. Therefore, potential future work is to include HD-Map as an optional input to our method and explored end-to-end motion forecasting. 

\mypar{Sensor Modalities.} Our current \name\ is a general query-based framework. Therefore, we could extend \name\ beyond the multi-camera setting and include LiDAR or Radar into our framework.

\section{Test Split Screenshot}
\label{sec:supp_test_screen_shot}

We provide the screenshot of the submission to nuScenes test set in Fig.~\ref{fig:screenshot}, as supplementary to the test set results in Tab.~\ref{tab:sota} (main paper).